\documentclass[letterpaper]{article} 
\usepackage{aaai2026}  
\usepackage{times}  
\usepackage{helvet}  
\usepackage{courier}  
\usepackage[hyphens]{url}  
\usepackage{graphicx} 
\usepackage{natbib}  
\usepackage{caption} 
\usepackage{algorithm}
\usepackage{algorithmic}
\usepackage{xspace}
\usepackage{amsfonts}
\usepackage{amsmath}
\usepackage{subcaption}

\newcommand{\goal}{\ensuremath{{\cal G}}\xspace}

\newcommand{\plan}{\ensuremath{\pi}\xspace}

\newcommand{\state}{\ensuremath{s}\xspace}

\newcommand{\action}{\ensuremath{a}\xspace}

\newcommand{\addeffects}{\ensuremath{\textsc{add}}\xspace}
\newcommand{\deleffects}{\ensuremath{\textsc{del}}\xspace}
\newcommand{\actionapplication}{\ensuremath{\gamma}\xspace}
\newcommand{\planapplication}{\ensuremath{\Gamma}\xspace}

\usepackage{newfloat}
\usepackage{xcolor}

\definecolor{lightgray}{rgb}{0.95,0.95,0.92}
\usepackage{listings}
\DeclareCaptionStyle{ruled}{labelfont=normalfont,labelsep=colon,strut=off} 
\floatstyle{ruled}
\newfloat{listing}{tb}{lst}{}

\floatname{listing}{Listing}
\usepackage{mathtools}
\nocopyright 

\title{Semantic Partial Grounding via LLMs}
\author{
    Giuseppe Canonaco\equalcontrib\textsuperscript{\rm 1},
    Alberto Pozanco\equalcontrib\textsuperscript{\rm 1},
    Daniel Borrajo\textsuperscript{\rm 1}
}
\affiliations{
    \textsuperscript{\rm 1}JPMorgan Chase \& Co. AI Research\\
    giuseppe.canonaco@jpmorgan.com, alberto.pozancolancho@jpmorgan.com, daniel.borrajo@jpmchase.com
}

\usepackage{bibentry}

\begin{document}

\maketitle

\begin{abstract}
Grounding is a critical step in classical planning, yet it often becomes a computational bottleneck due to the exponential growth in grounded actions and atoms as task size increases. Recent advances in partial grounding have addressed this challenge by incrementally grounding only the most promising operators, guided by predictive models. However, these approaches primarily rely on relational features or learned embeddings and do not leverage the textual and structural cues present in PDDL descriptions. We propose SPG‑LLM, which uses LLMs to analyze the domain and problem files to heuristically identify potentially irrelevant objects, actions, and predicates prior to grounding, significantly reducing the size of the grounded task. Across seven hard‑to‑ground benchmarks, SPG‑LLM achieves faster grounding—often by orders of magnitude—while delivering comparable or better plan costs in some domains.
\end{abstract}


\section{Introduction}

Domain-independent planning requires the world to be described using an appropriate representation language.
In classical planning, a common such language is the Planning Domain Definition Language (PDDL)~\cite{haslum2019introduction}, a first-order logic-based language designed to standardize research in planning. 

With some notable exceptions~\cite{correa2020lifted}, most current planners ground the first-order representation of the problem into a propositional encoding as a preprocessing step.
While this enables efficient implementation of heuristic and search algorithms, grounding can become impractical for large planning problems, as the size of the grounded task grows exponentially with the number of parameters in the predicates and actions of the planning domain.
In order to retain the advantages of grounded representations while mitigating the grounding bottleneck, recent literature has proposed addressing large planning instances through \emph{partial grounding}.
This technique involves grounding only the parts of the planning task that are deemed \emph{necessary} to find a plan, rather than grounding the entire task.
The approach is sound—if a plan is found for the partially grounded task, it is a valid plan for the original task—but incomplete, as the partially grounded task is only solvable if the operators in at least one plan have been grounded.

Most partial grounding approaches train predictive models that determine when an operator will be needed to solve a planning task.
In~\citet{gnad2019learning}, the authors use small planning tasks as a training set to extract relational rules as features, and then apply simple machine learning models to predict whether an operator should be grounded.
These predictive models are trained offline for each domain, relying only on the initial and goal descriptions to estimate the likelihood that a plan will include a given operator. 
In contrast,~\citet{areces2023partial} also utilize intermediate information from the relaxed planning graph and employ small language models (word embeddings) as predictors.
Instead of prioritizing certain operators during grounding,~\citet{silver2021planning} introduce a convolutional graph neural network architecture to predict the subset of objects sufficient for solving a planning problem.
This approach offers two main advantages over~\citet{gnad2019learning} and~\citet{areces2023partial}. 
First, by operating at the task level, it allows the planner's  grounding step to be treated as a black box, whereas the other methods require modifications to the planner’s grounding process. 
Second, by operating at the object level rather than over grounded action instances, it enjoys a computational advantage, as the object set is typically far smaller than the combinatorial space of grounded actions. 

In this paper, we also adopt a task-level approach and leverage recent advances in Large Language Models (LLMs) to identify not only potentially irrelevant objects, but also actions and predicates before grounding.
LLMs have been successfully used to prune the state space in robotic tasks~\cite{perez2025scalable}. 
We argue that many planning tasks specified in PDDL contain enough detail to provide rich semantic information, which LLMs can leverage to heuristically identify the sub-components of the task that are relevant for finding a plan.
Although this approach does not offer guarantees of completeness or soundness, it can be effective in practice for grounding and solving large planning tasks that possess semantic structure exposed in the PDDL files.

For instance, in the context of a cooking domain we may have a very deep and wide taxonomy of ingredients like: \textit{seafood}, \textit{meat}, \textit{vegetable}, and \textit{dairy}, with many objects of each type. We will have plenty of different actions like: \textit{chop}; \textit{slice}; \textit{boil}; \textit{mix}; and \textit{roast}. However, when it comes to finding the plan to accomplish a certain goal representing a recipe like cooking a \textit{Tagliatelle alla Bolognese}, we may take advantage of the common knowledge exposed in the goal and preemptively remove from the problem all the objects that are under the \textit{seafood} category, as well as all the \textit{roast} actions. This will result in less actions to be grounded because of the fewer amount of objects and action schemas present in the problem. 

The remainder of the paper is organized as follows. 
First, we formalize planning tasks and explain how they are described and grounded. 
Next, we present SPG-LLM, our semantic partial grounding approach designed to reduce the number of objects, predicates, and actions in a planning task. 
We then evaluate SPG-LLM against several baselines for grounding and solving planning tasks across different domains. 
Finally, we conclude by discussing the main findings and outlining potential directions for future research.

\section{Background}
A STRIPS \emph{planning task} is a tuple $\Pi = \langle {\cal P}, {\cal O}, {\cal A}, s_0, {\cal G} \rangle$, where $\cal P$ is a set of \emph{predicates}, $\cal O$ is a set of \emph{objects}, $\cal A$ is a set of \emph{action schemas}, $s_0$ is the \emph{initial state}, and $\cal G$ is a \emph{goal condition}.
Predicates $p \in {\cal P}$ have an associated arity.
If $p$ is an $n$-ary predicate and $\Vec{t}=\langle t_1, \ldots, t_n \rangle$ is a tuple of objects from $\cal O$, then $p(\Vec{t})$ is an \emph{atom}.
By \emph{grounding} the variables in an atom $p(\Vec{t})$—that is, substituting them with objects from $\cal O$—we obtain a \emph{ground atom}.
A \emph{state}, such as $s_0$, is the set of ground atoms that are true at that moment.
Similarly, the goal condition ${\cal G}$ is the set of ground atoms that must be true in a goal state.
An action schema $a[\Delta] \in {\cal A}$ is a tuple $\langle \textsc{pre}(a[\Delta]), \textsc{add}(a[\Delta]), \textsc{del}(a[\Delta]), \textsc{cost}(a[\Delta]) \rangle$, which specifies the \emph{precondition}, the \emph{add list}, \emph{delete list}, and \emph{cost} of $a[\Delta]$.
The cost is a non-negative real number $\mathbb{R}^0$, while the other three components are finite sets of atoms defined over $\cal P$.
Here, $\Delta$ refers to the set of free variables present in any atom within these components. 
By \emph{grounding} an action schema $a[\Delta]$—substituting its free variables $\Delta$ with objects from $\cal O$—we obtain a \emph{ground action} $a$, or simply an \emph{action}.
An action $a$ is \emph{applicable} in a state $s$ if $\textsc{pre}(a) \subseteq s$.
We define the result of applying an action in a state as $\actionapplication(\state,\action)=(\state \setminus \deleffects(\action)) \cup \addeffects(\action)$.
A sequence of actions $\plan=(\action_1,\ldots,\action_n)$ is applicable in a state $\state_0$ if there are states $(\state_1.\ldots,\state_n)$ such that $\action_i$ is applicable in $\state_{i-1}$ and $\state_i=\actionapplication(\state_{i-1},\action_i)$.
The resulting state after applying a \emph{sequence of actions} is $\planapplication(\state,\pi)=\state_n$, and $\textsc{cost}(\plan) = \sum_{i}^n \textsc{cost}(\action_i)$ denotes the cost of $\plan$. 
The solution to a planning task $\Pi$ is a \emph{plan}, i.e., a sequence of actions $\plan$ such that $\goal \subseteq \planapplication(s_0,\pi)$.

Planning tasks are described in PDDL~\cite{haslum2019introduction} by dividing the task definition into two files. The \emph{domain}, denoted as $D$, specifies the set of object types, the predicates $\cal P$, and the action schemas $\cal A$. The \emph{problem}, denoted as $P$, defines the set of objects $\cal O$ for the task, the initial state $s_0$, and the goal condition $\cal G$.

\section{Proposed Approach}\label{sec:approach}

In this section, we will focus on how to prune $\mathcal{P}$, $\mathcal{A}$, and $\cal O$ with the aim of speeding up the grounding phase, a notorious bottleneck for planning engines. We will do so by leveraging the semantics intrinsic to the domain and problem files.

\begin{algorithm}[!t]
	\caption{SPG-LLM}
	\label{alg:LLM-SP}
	\begin{algorithmic}[1]
        \STATE {\bfseries Input:} $D$, $P$, $\Phi$ the prompt template, and $K$ the maximum number of attempts
        \STATE {\bfseries Output:} $D'$, $P'$
        \STATE $\varphi$ = format($\Phi$, $D$, $P$)
        \STATE $D'$, $P'$ = LLM($\varphi$)
        \IF {validate$(D', P', D, P, \Phi, \varphi)$}
            \STATE \textbf{return} $D'$, $P'$
        \ENDIF
        \FOR{k=1 to K-1}
            \STATE $D'$, $P'$ = LLM($\varphi$)
            \IF {validate$(D', P', D, P, \Phi, \varphi)$}
                \STATE \textbf{return} $D'$, $P'$
            \ENDIF
        \ENDFOR
        \STATE \textbf{return} $\emptyset, \emptyset$
	\end{algorithmic}
\end{algorithm}

This semantic analysis is performed by an LLM, which leverages cues in the PDDL domain and problem descriptions to heuristically identify and propose pruning of objects, predicates, and actions that are unlikely to be required for the goal.
It is important to note that pruning an object from $\cal O$ requires a consistent update to the initial state, specifically by removing all atoms that involve the pruned object. 
This approach works at the task level, prior to grounding, hence it is complementary to the partial grounding solutions of~\citet{areces2023partial} and~\citet{gnad2019learning}, as well as to deterministic approaches to eliminate irrelevant facts and operators from grounded tasks~\cite{nebel1997ignoring,haslum2013safe,torralba2015focusing}.

In Algorithm~\ref{alg:LLM-SP}, we have reported the pseudo-code describing our solution called \textit{Semantic Partial Grounding via LLM} (SPG-LLM). In Line 3, we format the prompt to feed the LLM with the domain and problem descriptions. This is done by substituting the mentioned domain and problem descriptions placeholders in the prompt template with their actual descriptions via the \textit{format} function. Once the prompt is ready, in Line 4, we call the LLM. Soon after, in Line 5, we validate the pruned domain and problem. This validation is performed at 3 different levels: syntactic, computational, and semantic. Syntactic validation identifies errors related to output formatting and PDDL syntax compliance. 
Computational validation assesses whether a plan can be generated within a predefined time bound. This is done to verify that the reformulated task is solvable. 
Semantic validation verifies that the relationships $\mathcal{P}' \subseteq \mathcal{P}$, $\mathcal{A}' \subseteq \mathcal{A}$, $\cal O' \subseteq \cal O$ (this entails $s'_0 \subseteq s_0$), and $\mathcal{G}' \equiv \mathcal{G}$ are correctly maintained. 
This serves only as a soundness proxy to ensure that the LLM is not hallucinating objects or actions, rather than providing a soundness guarantee. Plans that solve the reformulated task may still not be valid solutions for the original task. 
If validation fails, the prompt template is populated with the original task description ($D$ and $P$), the LLM-generated output ($D'$ and $P'$), and a description of the identified error to facilitate correction in subsequent LLM interactions. The formatted prompt is stored into the variable $\varphi$ by the \textit{validate} function. If no errors are detected during validation, as indicated in Line 6, the process terminates. Otherwise, the procedure is retried up to $K-1$ additional times. The prompt template $\Phi$ employed to leverage common knowledge from LLMs for task simplification is presented in Appendix~\ref{appx:prompt}.

\section{Evaluation}
\subsection{Experimental Setting}

\paragraph{Benchmark.} To evaluate the performance of SPG-LLM, we selected a diverse set of planning domains: \textit{Agricola}, \textit{Blocksworld}, \textit{Depots}, \textit{Hiking}, \textit{Satellite}, \textit{TPP}, and \textit{Zenotravel}. These domains are the same as those used by~\citet{areces2023partial}. For each domain, we evaluate the various approaches on a fixed set of 25 randomly selected tasks from their evaluation set. This evaluation set comprises hard to ground planning problems specifically designed to assess the effectiveness of partial grounding techniques. 
We chose this benchmark over other hard-to-ground benchmarks~\cite{lauer2021polynomial} because we wanted the tasks to be challenging, yet sufficiently small to allow grounding and solving within reasonable time and memory limits using standard planners. This selection enables us to assess not only the grounding capabilities of SPG-LLM, but also how its task reductions impact the performance of state-of-the-art planners when solving these tasks.

\paragraph{Approaches.} We compare SPG-LLM against Full Grounding (FG)~\cite{helmert2009concise} and Planning with Learned Object Importance (PLOI)~\cite{silver2021planning}. FG serves as a simple baseline, instantiating all facts and operators that are delete-free reachable from the initial state~\cite{hoffmann2003metric}. PLOI leverages a graph neural network to prioritize the objects to be included in the problem. The approach of~\citet{silver2021planning} was selected over those proposed by~\citet{gnad2019learning} and~\citet{areces2023partial}, as it operates at the task level, thus being the most closely related technique to our work in the  literature.  PLOI was trained with the following parameters over $3$ random seeds per domain: $200$ iterations, $120$ seconds timeout, $100$ training problems, with incremental planning to generate the proposed pruned problem, and the first iteration of LAMA~\cite{richter2010lama} to collect training data. We restrict our evaluation of PLOI to domains containing predicates of arity two or less, reflecting a current limitation of the method. SPG-LLM has been run with $K=1$ and \textit{GPT-5-2025-08-07} (\textit{verbosity="medium"} and \textit{reasoning\_effort="medium"}) as LLM model.

\paragraph{Metrics.} We report the following metrics, where lower values indicate better performance: grounding time (in seconds) and the number of grounded actions; solving time (in seconds), and the plan cost for the best plan found by LAMA within the specified limits. Using optimal planners was not feasible in the context of these problems due to their size.
We verify that all the plans generated by solving the different reformulated tasks are sound wrt. the original task by using VAL~\cite{howey2004val}. 

\paragraph{Reproducibility.} SPG-LLM and PLOI have been run on mac OS Sequoia using an 8 cores Apple M2 processor and 24 GB of RAM to generate the reformulated (pruned) tasks.
All the tasks have then been solved running LAMA  on an Intel Xeon E5-2666 v3 CPU @ 2.90GHz x 8 processors with a 8GB memory bound and a time limit of 1800s.

\begin{table*}[]
\scriptsize
\setlength{\tabcolsep}{0.5pt}
    \centering
    \resizebox{\linewidth}{!}{
    \begin{tabular}{|l|r|r|r||r|r|r||r|r|r||r|r|r|}
    \hline
& \multicolumn{3}{c||}{Grounded Actions} & \multicolumn{3}{c||}{Grounding Time} &  \multicolumn{3}{c||}{Plan Cost} & \multicolumn{3}{c|}{Solving Time} \\ \hline
        Domain (\#Tasks) & \multicolumn{1}{c|}{FG} & \multicolumn{1}{|c|}{PLOI} & \multicolumn{1}{c||}{SPG}  & \multicolumn{1}{c|}{FG} & \multicolumn{1}{|c|}{PLOI} & \multicolumn{1}{c||}{SPG} & \multicolumn{1}{c|}{FG} & \multicolumn{1}{|c|}{PLOI} & \multicolumn{1}{c||}{SPG} & \multicolumn{1}{c|}{FG} & \multicolumn{1}{|c|}{PLOI} & \multicolumn{1}{c|}{SPG} \\ \hline
        Agricola (8)& $678k \pm 324k$ &- &$\boldsymbol{146k \pm 329k}$ &$103.0 \pm 51.2$ &- &$\boldsymbol{22.0 \pm 51.5}$ &$7752.1 \pm 2216.7$ &- &$7752.5 \pm 2217.4$ &$\boldsymbol{156.7 \pm 84.3}$ &- &$251.3 \pm 555.3$ \\
        
        Blocksworld (24) & $72k \pm 5k$ &$72k \pm 5k$ &$\boldsymbol{46k \pm 34k}$ &$7.8 \pm 0.6$ &$7.7 \pm 0.6$ &$\boldsymbol{4.8 \pm 3.6}$ &$94.9 \pm 17.4$ &$95.2 \pm 17.3$ &$\boldsymbol{82.5 \pm 10.5}$ &$704.7 \pm 522.2$ &$693.2 \pm 507.3$ &$\boldsymbol{396.2 \pm 516.1}$\\
        
        Depots (24) & $16k \pm 1k$ &$11k \pm 2k$ &$\boldsymbol{6k \pm 2k}$ &$2.2 \pm 0.3$ &$1.5 \pm 0.3$ &$\boldsymbol{0.8 \pm 0.3}$ &$74.1 \pm 15.3$ &$71.6 \pm 14.7$ &$\boldsymbol{66.5 \pm 11.3}$ &$192.0 \pm 349.2$ &$338.1 \pm 472.0$ &$\boldsymbol{158.2 \pm 271.2}$    \\
        
        Hiking (15) & $420k \pm 263k$ &- &$\boldsymbol{69k \pm 116k}$ &$67.3 \pm 43.8$ &- &$\boldsymbol{10.1 \pm 18.2}$ &$\boldsymbol{69.9 \pm 5.1}$ &- &$76.4 \pm 12.0$ &$\boldsymbol{288.8 \pm 438.7}$ &- &$337.0 \pm 445.4$ \\
        
        Satellite (17) & $327k \pm 262k$ &$142k \pm 224k$ &$\boldsymbol{91k \pm 135k}$ &$20.4 \pm 16.2$ &$\boldsymbol{4.7 \pm 3.7}$ &$5.3 \pm 8.1$ &$\boldsymbol{283.2 \pm 124.6}$ &$2220.3 \pm 5548.4$ &$291.8 \pm 123.5$ &$244.5 \pm 358.0$ &$\boldsymbol{136.3 \pm 220.9}$ &$179.3 \pm 410.1$ \\
        
        TPP (22) & $385k \pm 199k$ &- &$\boldsymbol{17k \pm 9k}$ &$42.0 \pm 21.6$ &- &$\boldsymbol{1.9 \pm 0.9}$ &$\boldsymbol{623.9 \pm 124.3}$ &- &$850.0 \pm 168.0$ &$695.7 \pm 440.8$ &- &$\boldsymbol{27.8 \pm 40.1}$ \\
        
        Zenotravel (12) & $243k \pm 110k$ &$377k \pm 506k$ &$\boldsymbol{48k \pm 17k}$ &$19.1 \pm 8.9$ &$8.4 \pm 2.8$ &$\boldsymbol{3.1 \pm 1.2}$ &$\boldsymbol{457.0 \pm 91.1}$ &$7389.6 \pm 13090.9$ &$493.2 \pm 102.5$ &$1154.2 \pm 503.8$ &$719.0 \pm 660.0$ &$\boldsymbol{371.3 \pm 353.1}$ \\
         \hline
    \end{tabular}}
    \caption{Results of the three approaches across commonly solved tasks in the benchmark. Best performance in bold.}
    \label{tab:big_table}
\end{table*}

\subsection{Results}

\begin{figure*}
    \centering
    \begin{subfigure}[t]{0.24\linewidth}\centering
    \includegraphics[width=\textwidth]{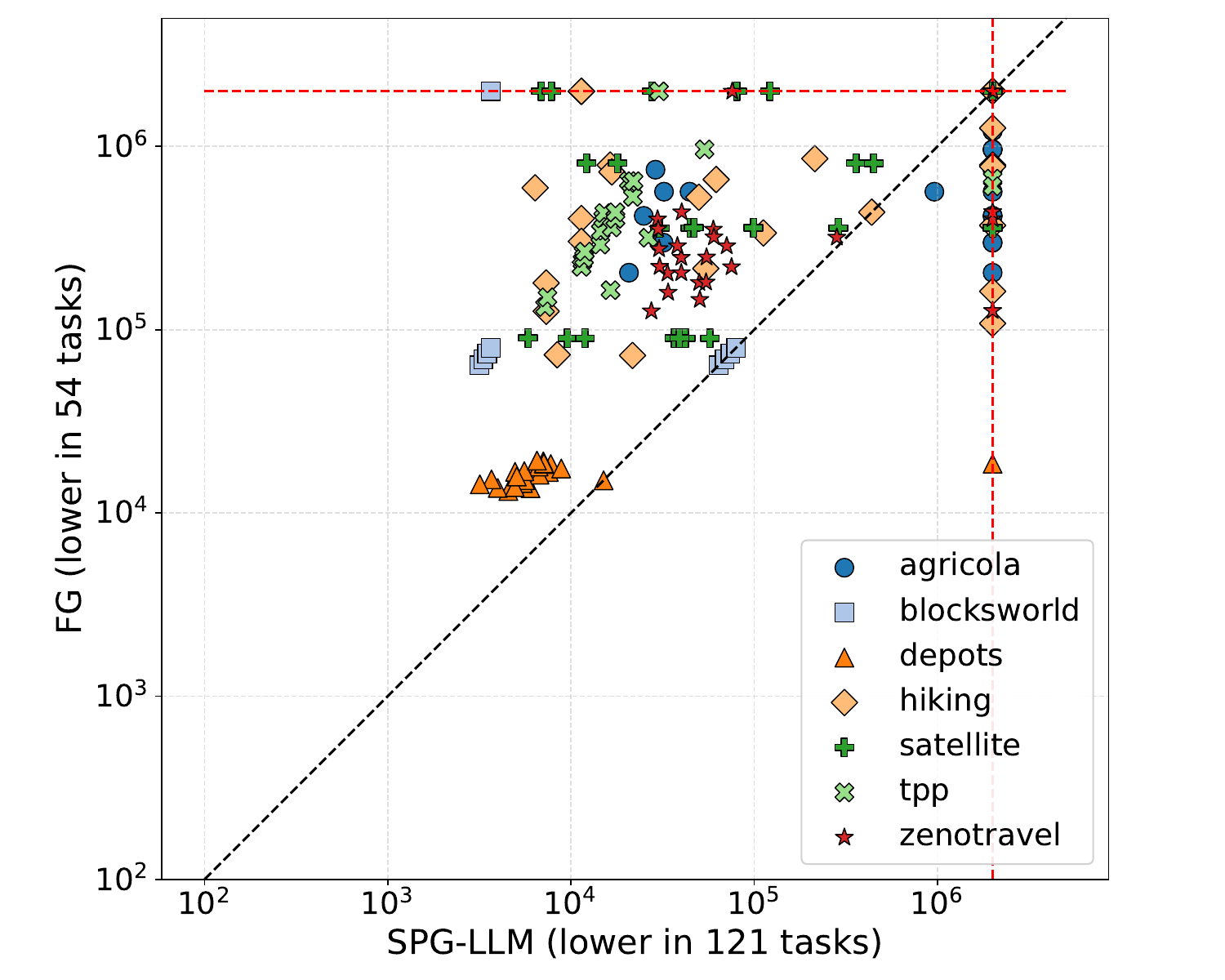}
    \caption{SPG vs FG actions.}
    \label{fig:fg_vs_spg_ops}
    \end{subfigure}
    \begin{subfigure}[t]{0.24\linewidth}\centering
    \includegraphics[width=\textwidth]{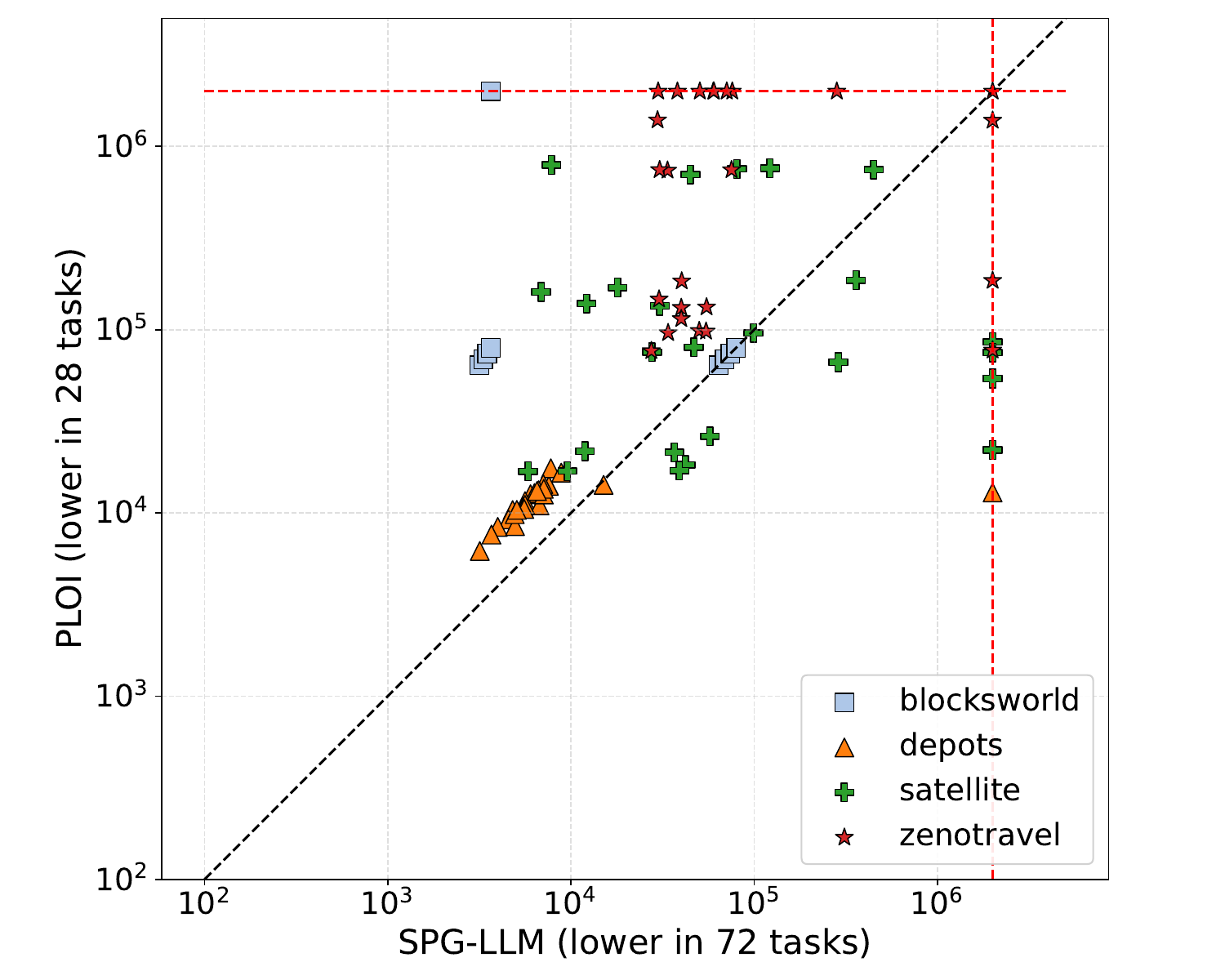}
    \caption{SPG vs PLOI actions.}
    \label{fig:ploi_vs_spg_ops}
    \end{subfigure}
    \hfill
     \begin{subfigure}[t]{0.24\linewidth}\centering
    \includegraphics[width=\textwidth]{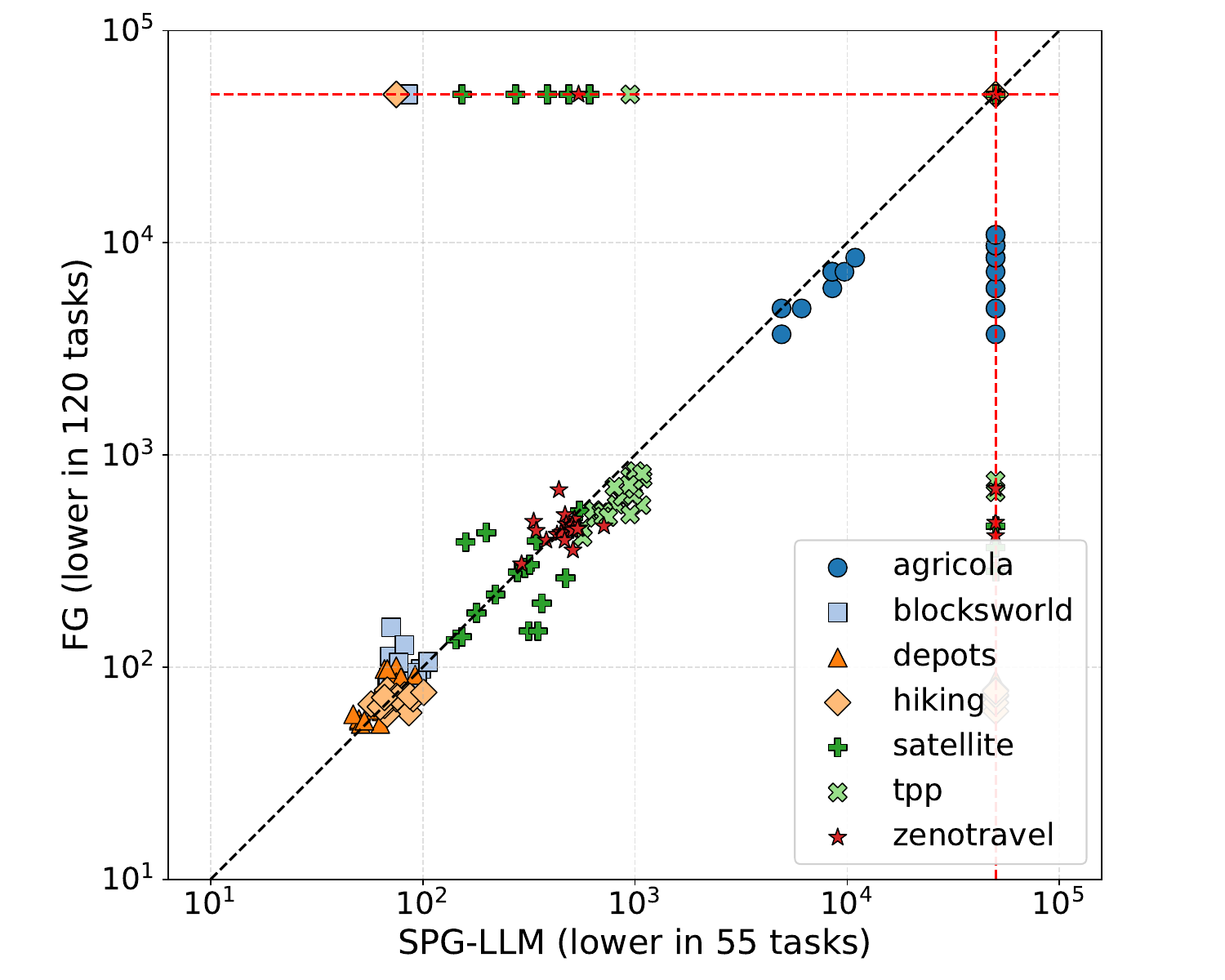}
    \caption{SPG vs FG plan cost.}
    \label{fig:fg_vs_spg_cost}
    \end{subfigure}
    \hfill
     \begin{subfigure}[t]{0.24\linewidth}\centering
    \includegraphics[width=\textwidth]{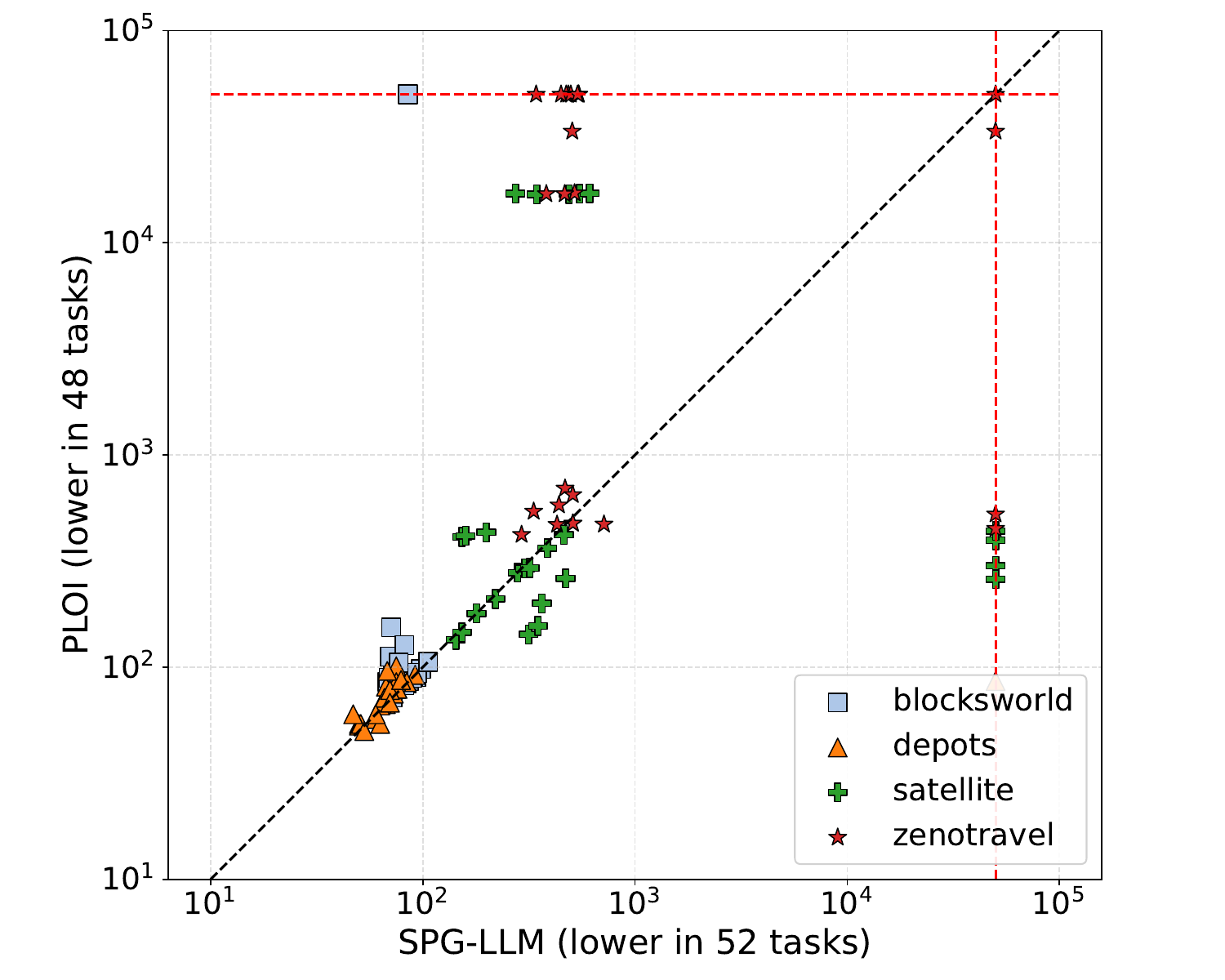}
    \caption{SPG vs PLOI plan cost.}
    \label{fig:ploi_vs_spg_cost}
    \end{subfigure}
    \hfill
    \caption{Number of grounded actions (Figures \ref{fig:fg_vs_spg_ops} and \ref{fig:ploi_vs_spg_ops}) and plan cost (Figures \ref{fig:fg_vs_spg_cost} and \ref{fig:ploi_vs_spg_cost}) produced by each baseline ($y$-axis) and SPG-LLM ($x$-axis). Each point is a problem, with markers indicating the domain. Points above the diagonal indicate SPG-LLM have better performance. Points within the red dashed lines indicate cases where the approaches either failed to solve the grounded task within the time and memory limits, or produced a grounding that resulted in a plan not valid for the original task.}
    \label{fig:analysis}
\end{figure*}

\paragraph{Grounding Results.} Our first analysis aim to understand whether SPG-LLM improves grounding by generating a lower number of operators and reducing the grounding time.
Figures \ref{fig:fg_vs_spg_ops} and \ref{fig:ploi_vs_spg_ops} present the number of actions grounded by SPG-LLM in tasks that are both solvable and sound, in comparison to FG and PLOI, respectively.
SPG-LLM outperforms both baselines by grounding fewer operators in most tasks, achieving reductions of up to two orders of magnitude. 
Although consistent across domains, we can identify some peculiar patterns that are worth discussing. In Blocksworld, we may see two groups of results: one of them lying on the bisector, and the other above it. This is due to the fact that SPG-LLM leaves all the problem files unchanged and removes a superfluous action, \textit{move-b-to-b}, from the domain in $10$ out of $25$ tasks. This action schema moves one clear block onto another, but it is superfluous because the same effect can be achieved using the other two action schemas: first moving clear blocks to the table, and then from the table onto clear blocks. Another pattern that can be clearly noticed is the fact that, in Depots, SPG-LLM is able to provide a smaller improvement w.r.t. other domains like Satellite. This is due to the fact that in Depots our approach can remove a smaller proportion of objects w.r.t. Satellite where there are more irrelevant objects. 
Overall, the reduction in grounded operators achieved by SPG-LLM is reflected in the grounding times shown in Table~\ref{tab:big_table}, where SPG-LLM is the fastest approach across the commonly solved tasks in all domains.
These benefits come with a trade-off, as SPG-LLM produces 36 tasks out of the 175 in the benchmark that cannot be solved to yield valid plans for the original task. 
The number of invalid tasks varies by domain; for example, in Agricola, SPG-LLM generates only 8 tasks where sound plans can be computed, while in Blocksworld, all tasks grounded by SPG-LLM are valid.

\paragraph{Solving Results.} Our second analysis aims to asses the impact on performance the reformulated tasks have compared to solving the original task.
Regarding coverage, FG is able to generate valid plans for the original task within the time and memory limits in 161 out of 175 tasks, compared to 139 out of 175 tasks for SPG-LLM, and 90 out of 100 tasks for PLOI.
Figures \ref{fig:fg_vs_spg_cost} and \ref{fig:ploi_vs_spg_cost} show the lowest plan cost achieved by LAMA when solving tasks generated by SPG-LLM, compared to those generated by FG and PLOI, respectively. 
In this case, the results vary depending on the domain. In domains such as Agricola, Satellite, and Zenotravel, FG and SPG-LLM produce plans of similar quality. 
However, in Agricola, SPG-LLM’s grounding appears to negatively impact search time, while search times are comparable in Satellite and significantly faster for SPG-LLM’s reformulated tasks in Zenotravel. 
In Blocksworld and Depots, SPG-LLM grounding leads to both better plans and shorter solving times. 
SPG-LLM enables the generation of lower-cost plans because, with a smaller task, LAMA can perform more iterations using algorithms that may find solutions closer to optimal.
Finally, in Hiking and TPP, FG yields lower-cost plans, but finding them in Hiking is faster, and in TPP, solving the SPG-LLM grounding can be up to two orders of magnitude faster.
It is also noteworthy that SPG-LLM grounding enables LAMA to find plans for 10 tasks that cannot be solved using the standard FG grounding.

Taking a more qualitative perspective, we analyzed problems in Zenotravel and found that our approach identifies the \textit{zoom} action schema as irrelevant, since it is simply a faster variant of \textit{fly} that consumes more fuel. 
Additionally, SPG-LLM simplifies the problems by removing objects.
For one of the instances, it removes 3 persons, namely \textit{person 5}, \textit{person 14}, and \textit{person 33} that are not involved in the goal, hence they do not need to be transported. 
It also reduces the number of aircraft objects from 24 to 11, and the fuel levels from 7 to 2.
All these changes may result in longer plans because fewer resources are available, requiring actions like \textit{refuel} to be used more frequently. 
However, this grounding in Zenotravel enables, on average, a reduction of one order of magnitude in both grounding and solving time, while only increasing the average plan cost from $457.0$ to $493.2$.

\section{Conclusions and Future Work}
In this paper, we introduced SPG‑LLM, a pre‑grounding, task‑level approach that uses an LLM to analyze the PDDL domain and problem files and heuristically prune objects, action schemas, and predicates, followed by syntactic, computational, and semantic checks. Across seven hard‑to‑ground benchmarks, SPG‑LLM consistently reduces the number of grounded operators and achieves the fastest grounding times relative to FG and PLOI on commonly solved tasks. This happens with domain‑dependent trade‑offs in plan cost and overall coverage.

In future work we would like to enhance SPG-LLM by improving the validation checks, incorporating soundness checks for a subset of the plans allowed in the reformulated task.
This should lead to an increase in the number of reformulated tasks for which planners can generate plans that are valid in the original task.
We also plan to investigate the generation of explanations~\cite{chakraborti2020emerging} based on the task reformulations suggested by SPG-LLM, to better understand the reasoning behind pruning specific objects or actions.
Finally, we would like to experiment with real-world domains that have richer semantics, such as the one described in the introduction. 
In these domains, we anticipate even greater performance improvements.
\section*{Disclaimer}
This paper was prepared for informational purposes  by the Artificial Intelligence Research group of JPMorgan Chase \& Co. and its affiliates ("JP Morgan'') and is not a product of the Research Department of JP Morgan. JP Morgan makes no representation and warranty whatsoever and disclaims all liability, for the completeness, accuracy or reliability of the information contained herein. This document is not intended as investment research or investment advice, or a recommendation, offer or solicitation for the purchase or sale of any security, financial instrument, financial product or service, or to be used in any way for evaluating the merits of participating in any transaction, and shall not constitute a solicitation under any jurisdiction or to any person, if such solicitation under such jurisdiction or to such person would be unlawful.

© 2024 JPMorgan Chase \& Co. All rights reserved.
\bibliography{bibliography}

\clearpage
\appendix
\section{Prompt Template}\label{appx:prompt}
The template prompt $\Phi$, presented in Listing~\ref{lst:prompt-template}, is structured in the following way. The first part instructs the LLM about what it has to do. Then we provide an example to further clarify the task it has to accomplish. The example is removed for conciseness. However, in a nutshell, it reports an example of a domain and problem we want to prune. The pruned domain with the rationale that allowed the pruning. The pruned problem with what we kept and the reason why we kept it, and analogously for what we removed. Finally, the domain and problem the LLM is supposed to work on.
\begin{listing}[!h]%
\caption{Prompt Template}%
\label{lst:prompt-template}%
\begin{lstlisting}[]
You are a PDDL expert and you will help me out simplifying a planning task according to the goal we have to achieve in the planning task itself. The simplification will revolve around objects removal from the problem.pddl file, and actions or predicates from the domain.pddl file. The objects removal can be done analyzing the hieararchy of objects types and then selecting the objects (or types of objects) to be removed because not necessary to achieve the goal. The predicates that can be removed are as well the ones that are not necessary to achieve the goal. Be careful about the interplay between actions, predicates, and goal to understand what can be removed and what not.

The goal cannot be modified. So all the objects and predicates present in the goal cannot be removed.
Let me further elucidate with an example:
Domain
<Removed for conciseness>
Original Problem
<Removed for conciseness>
New Domain:
<Removed for conciseness>
Reason:
<Removed for conciseness>
Simplified Problem:
<Removed for conciseness>
Removed:
<Removed for conciseness>
Kept:
<Removed for conciseness>
This is the domain we are going to operate on:
<Domain.pddl>
This is the problem:
<Problem.pddl>
Please output only the modified files ready to be fed to a planner.
\end{lstlisting}
\end{listing}

\end{document}